%% file: ictai_yuv_2026_arxiv.tex
\documentclass[conference]{IEEEtran}
\IEEEoverridecommandlockouts

\usepackage[numbers,sort&compress]{natbib}
\usepackage{amsmath,amssymb,amsfonts}
\usepackage{graphicx}
\usepackage{textcomp}
\usepackage{xcolor}

\usepackage{microtype}
\usepackage{hyperref}
\usepackage{url}
\usepackage{booktabs}
\usepackage{threeparttable}
\usepackage{algorithm}
\usepackage{algpseudocode}
\makeatletter
\newcommand\fs@compactruled{%
  \fs@ruled
  \def\@fs@post{\kern2pt\hrule\relax\vskip-4pt}}
\makeatother
\floatstyle{compactruled}
\restylefloat{algorithm}
\usepackage{makecell}
\usepackage{multirow}
\usepackage{enumitem}
\usepackage{verbatim}
\usepackage{tabularx}
\usepackage{adjustbox}

\setlist[itemize]{leftmargin=*,topsep=2pt,itemsep=1pt,parsep=0pt,partopsep=0pt}

\usepackage{lineno}
\usepackage{todonotes}
\usepackage{color}

\definecolor{darkblue}{rgb}{0, 0, 0.5}
\hypersetup{colorlinks=true, citecolor=darkblue, linkcolor=darkblue, urlcolor=darkblue}

\newcommand{\ours}{\textsc{ClosedLoopGaussian}}      
\newcommand{\Acc}{\mathbf{A}}                         

\newcommand{\clip}{\operatorname{clip}}
\newcommand{\pos}[1]{\left(#1\right)_{+}}             

\def\BibTeX{{\rm B\kern-.05em{\sc i\kern-.025em b}\kern-.08em
    T\kern-.1667em\lower.7ex\hbox{E}\kern-.125emX}}

\begin{document}

\title{{\small Accepted for publication in the 2026 IEEE 38th International Conference on Tools with Artificial Intelligence (ICTAI).}\\[0.5em]
  When Does Closed-Loop Curriculum Adaptation Help GRPO? An Analysis on Countdown}

 \author{\IEEEauthorblockN{Yuv Raj Pant\textsuperscript{$\ast$}, Moshiur Rahman Tonmoy\textsuperscript{$\dag$}, and  Kaiqun Fu\textsuperscript{$\ast$}}
 \IEEEauthorblockA{\textit{\textsuperscript{$\ast$}Department of Computer Science, Texas Christian University} \\
 \textit{\textsuperscript{$\dag$}Department of Electrical Engineering and Computer Science, Texas A\&M University - Kingsville}\\
 \{yuv.raj.pant, k.fu\}@tcu.edu, moshiur\_rahman.tonmoy@students.tamuk.edu}
 }

\maketitle

\begingroup
\renewcommand{\thefootnote}{}
\renewcommand{\theHfootnote}{ieeecopyright}
\footnote{© 2026 IEEE. Personal use of this material is permitted. Permission from IEEE must be obtained for all other uses, in any current or future media, including reprinting/republishing this material for advertising or promotional purposes, creating new collective works, for resale or redistribution to servers or lists, or reuse of any copyrighted component of this work in other works.}
\addtocounter{footnote}{-1}
\endgroup

\begin{abstract}
Reinforcement fine-tuning has become an effective post-training paradigm for
improving the reasoning ability of large language models, but it remains
sample-inefficient when the training distribution is mismatched to the
model's current capability. In group-relative objectives such as Group Relative Policy Optimization (GRPO),
problems that are too hard often produce uniformly failed rollouts, while
problems that are too easy provide little useful advantage signal. In this
work, we introduce \textsc{ClosedLoopGaussian}, a lightweight adaptive
curriculum method for efficient reinforcement fine-tuning on ordered
reasoning tasks. Instead of following a fixed easy-to-hard schedule, \textsc
{ClosedLoopGaussian} preserves the structure of an Easy-to-Hard Gaussian
sampler while dynamically adjusting its pace and spread using periodic probe
estimates of per-difficulty accuracy. The controller accelerates training
when the model is ahead of a reference curriculum, slows down when the model
falls behind, and broadens sampling when earlier difficulties regress. This
feedback mechanism keeps training focused on tasks that are challenging but
still learnable, without changing the reward function, model architecture, or
underlying GRPO optimizer. Experiments on symbolic arithmetic reasoning task
show that \textsc{ClosedLoopGaussian} improves exact-match accuracy over
vanilla GRPO and fixed open-loop curricula, with the strongest gains in
intermediate-difficulty regimes where group-relative learning signals are
most informative. Codes are available at
\url{https://github.com/yuvrajpant56/Closed_Loop_Curriculum}.
\end{abstract}


\input{sections/introduction}
\input{sections/method}
\input{sections/experiment}
\input{sections/limitations}
\input{sections/conclusion}
\input{sections/acknowledgment}

\bibliography{references}
\bibliographystyle{ieeetr}

\end{document}

%% file: sections/introduction.tex
\section{Introduction}

Reinforcement Learning with Verifiable Rewards (RLVR) \citep {guo2025deepseekr1,
kimi2025k15, jaech2024openaio1} has become an important post-training paradigm
for improving the reasoning behavior of large language models (LLMs). Instead of
imitating fixed solution traces, an RLVR system samples candidate solutions,
scores them with a verifier such as exact-answer matching or executable checks,
and updates the policy toward higher-reward responses, improving performance on
mathematical, programming, and symbolic reasoning tasks \citep
{guo2025deepseekr1,jaech2024openaio1}. Group-relative methods such as GRPO
further make RLVR practical by estimating advantages from multiple sampled
responses to the same prompt, avoiding the need for a separately trained critic
\citep{shao2024deepseekmath}, and have been applied beyond mathematical
reasoning, for example to train generative judges
\citep{chen2025generativejudge}.

\begin{figure}[t]
    \centering
    \includegraphics[width=\columnwidth]{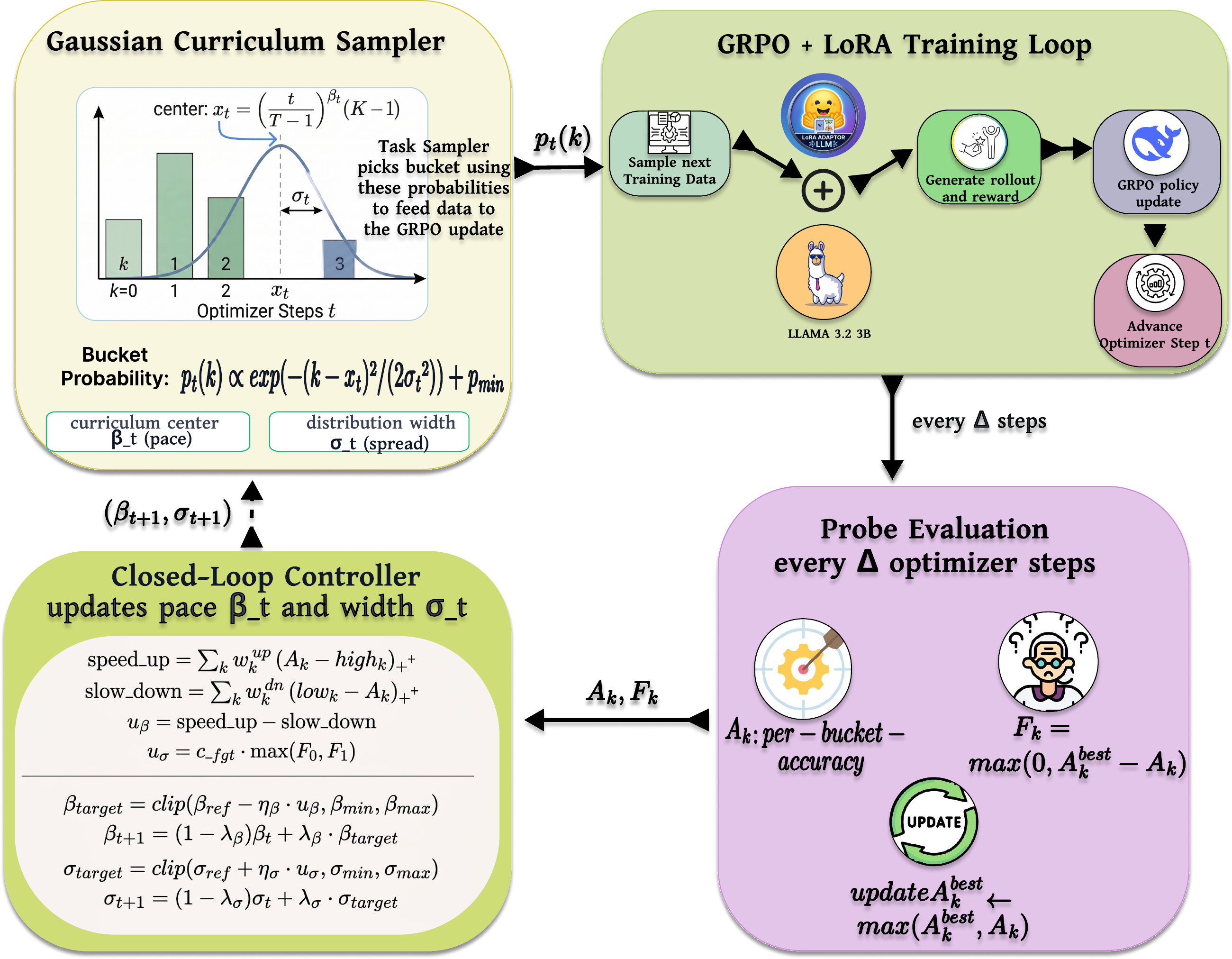}
    \vspace{-10pt}
    \caption{Overview of the proposed adaptive closed-loop curriculum framework.}
    \label{fig: adaptive_curriculum_overview}
    \vspace{-10pt}
\end{figure}

Despite this progress, GRPO remains highly sensitive to the difficulty of the
problems selected for training. When a problem is too hard for the current
policy, most rollouts fail, producing sparse rewards and little useful advantage
signal \citep{yu2025dapo}. When a problem is already too easy, sampled responses
become uniformly correct, again reducing within-group reward variation. In both
regimes the expensive rollout budget yields a weak gradient; indeed, Bae et al.
\citep{bae2025onlinedifficulty} show formally that expected policy improvement
is maximized on tasks of intermediate difficulty, where success probabilities
have the highest variance. A family of methods therefore modifies the
optimization or rollout procedure itself: DAPO filters and resamples prompts
until each batch contains groups with non-zero reward variance
\citep{yu2025dapo}, GSPO changes the granularity of policy optimization
\citep{zheng2025gspo}, and ShareGRPO improves the construction and reuse of
group-relative learning signals \citep{yao2025sharegrpo}. These approaches
improve how learning signals are extracted from sampled responses, but online
filtering spends variable rollout cost per update.

Curriculum learning addresses difficulty mismatch by organizing training from
easier to harder tasks as competence develops \citep{bengio2009curriculum}, and
automatic curriculum methods select tasks from learning progress, competence
estimates, or a learned teacher policy \citep
{graves2017automated,matiisen2020teacher,narvekar2020curriculum}.
In reinforcement fine-tuning for reasoning, Parashar et al.
\citep{parashar2026e2hreasoner} propose probabilistic easy-to-hard schedulers,
including a Gaussian schedule (E2H-G) that smoothly shifts training mass across
ordered difficulty buckets. These schedules are \emph{open-loop}: their pace and
spread are fixed before training and followed regardless of the model's actual
learning trajectory. This is fragile when progress varies across seeds and
difficulty levels; a model may learn faster than expected, stall on intermediate
tasks, or regress on earlier skills while the curriculum continues to advance.
Closer to our setting, a recent line of adaptive schedulers closes this gap from
observed training signals: ADaRFT(Adaptive Curriculum Reinforcement Finetuning)
steers a scalar target difficulty from recent rewards \citep{shi2025adarft},
DUMP schedules sampling across data distributions with a bandit rule over
advantage magnitudes \citep{wang2025dump}, and SEC(Self-Evolving Curriculum)
learn a task-selection policy alongside training \citep{chen2025selfevolving}.
These methods demonstrate the value of matching training tasks to current
capability, but they generally represent curriculum state through either a
task-selection policy or a single target-difficulty variable, which does not
explicitly separate two distinct curriculum decisions: how quickly to advance
toward harder tasks, and how broadly to rehearse previously encountered
difficulties. 

The central question of this paper is: \emph{Can a curriculum scheduler use
{feedback} from the model's own learning trajectory to synchronize task
selection with current capability, and does this closed-loop control improve
reinforcement fine-tuning over a {fixed easy-to-hard schedule}?} We answer this
question with \ours{}, a closed-loop curriculum scheduler for GRPO-style
reasoning post-training. 
Particularly, the paper makes the following three key contributions:
\begin{itemize}
\item \textbf{Algorithm.} \ours{}, a closed-loop scheduler that
 adapts the pace $\beta_t$ and spread $\sigma_t$ of a Gaussian easy-to-hard
 sampler from per-difficulty pass rates.

\item \textbf{Analysis.} Formalizing curriculum scheduling for
 group-relative reinforcement learning as a compute-allocation problem,
 decoupling control of rehearsal breadth.

\item \textbf{Evaluation.} A four-seed comparison against vanilla GRPO and a
 fixed open-loop curriculum on Countdown,  
 under an identical rollout budget, showing improved average accuracy, where group-relative
 learning signals are most informative.
\end{itemize}

%% file: sections/method.tex
\section{Methodology}
\label{sec:method}

\ours{} is a closed-loop curriculum scheduler for GRPO-style reasoning post-training that changes only the distribution from which training prompts are sampled, while leaving the underlying model, verifier, reward function, and optimizer unchanged. At each step, a Gaussian sampler assigns probabilities over ordered difficulty buckets, while periodic probe evaluations estimate bucket-wise accuracy. A feedback controller then adjusts the curriculum pace $\beta_t$ and spread $\sigma_t$ so that training can progress toward harder problems, slow when performance lags, or broaden rehearsal when earlier difficulties regress. A single scheduler object is shared by the sampler, trainer, and probe evaluator, so updates to $\beta_t$ or $\sigma_t$ affect the next sampled batch. From a compute-allocation perspective, the sampler determines how the fixed
rollout budget is distributed across difficulty levels. The controller therefore
adapts $\beta_t$ and $\sigma_t$ to reallocate training compute according to the
model's observed competence. Figure~\ref{fig: adaptive_curriculum_overview} summarizes the pipeline. The remainder of this section describes the difficulty buckets (\S\ref{sec:difficulty_buckets}), Gaussian sampler (\S\ref{sec:gaussian_sampler}), and probe-driven closed-loop controller (\S\ref{sec:closed_loop_controller}).

\subsection{Ordered Difficulty Space}
\label{sec:difficulty_buckets}

Each Countdown instance provides a set of input numbers and a target integer,
and requires an arithmetic expression over $+$, $-$, $\times$, $\div$ that
uses each input number exactly once and evaluates to the target. The data
provide fixed ordered difficulty groups indexed by $n\in\{2,3,4,5\}$, the
number of input operands in the instance; larger $n$ increases the
combinatorial search space and is empirically harder for the policy. We denote
these preassigned groups by $\mathcal{D}_0,\ldots,\mathcal{D}_{K-1}$. The
curriculum does not infer or relabel difficulty: at each step it samples a
bucket $k\sim p_t$ and then samples an example uniformly from $\mathcal{D}_k$.
Exact-answer correctness is determined independently by the unchanged
Countdown verifier.

\subsection{Gaussian Curriculum Sampler}
\label{sec:gaussian_sampler}

The sampler follows the same Gaussian easy-to-hard form as the fixed E2H-G
baseline~\citep{parashar2026e2hreasoner}, which keeps the sampling family
fixed and makes feedback the primary design difference between the two arms. At optimizer step $t$ with total optimizer budget $T$, the curriculum
center is
\begin{equation}
    x_t \;=\; \left(\frac{t}{T-1}\right)^{\beta_t}\,(K-1)
    \label{eq:adaptive_center}
\end{equation}
and bucket $k$ is sampled with probability
\begin{equation}
    p_t(k) \;=\;
    \frac{\exp\!\left(-\dfrac{(k-x_t)^2}{2\sigma_t^2}\right) + p_{\min}}
         {\displaystyle\sum_{j=0}^{K-1}
          \left[\exp\!\left(-\dfrac{(j-x_t)^2}{2\sigma_t^2}\right) + p_{\min}\right]}
    \label{eq:adaptive_probs}
\end{equation}
The pace $\beta_t$ controls how quickly $x_t$ migrates from the easiest bucket
toward the hardest: smaller $\beta_t$ advances the curriculum sooner, larger
$\beta_t$ holds it on easier material longer. The spread $\sigma_t$ controls
rehearsal breadth: larger $\sigma_t$ flattens the distribution so neighboring
buckets stay in the rollout mix. The additive floor $p_{\min}>0$ is applied
to each unnormalized weight \emph{before} normalization, guaranteeing every
bucket retains nonzero probability; this keeps the hardest bucket from being
starved entirely early in training while its share stays negligible until the
controller deems it worthwhile. Unlike \eqref{eq:adaptive_center}--\eqref{eq:adaptive_probs} with $\beta_t$ and
$\sigma_t$ held constant at predefined reference values for all $t$; the
proposed method instead adapts these parameters over time.

\subsection{Closed-Loop Curriculum Controller}
\label{sec:closed_loop_controller}

The controller closes the loop between the curriculum clock and the model's observed competence using a fixed, non-learned feedback rule based on two inexpensive, interpretable signals. It remains anchored to the reference trajectory, reverting to the open-loop baseline when no corrective signal is present, and can be viewed as an RLVR-specific form of learning-progress-based automatic curriculum learning \citep{graves2017automated,narvekar2020curriculum}.

\paragraph{Probe evaluation and calibration}
Every $\Delta=100$ optimizer steps, the controller greedily evaluates a fixed, pre-cached set of $100$ held-out examples per difficulty bucket using the unchanged Countdown verifier, producing $\Acc=(A_0,\ldots,A_{K-1})$. This probe is the only additional GPU computation introduced by \ours{}; indicative timing is reported in \S\ref{sec:results}. Controller parameters were coarsely calibrated to keep feedback responsive but bounded: $h$ and $\ell$ were chosen from reference mid-training pass rates, while gains and clipping ranges were set conservatively to limit deviation from the E2H-G reference. Settings were then frozen before the reported comparison; seed 42 was used during calibration.

\paragraph{Pace control via a symmetric, calibrated signal}
For each bucket, high and low thresholds $h_k$ and $\ell_k$ ($\ell_k<h_k$) are calibrated against mid-training pass rates, so $A_k>h_k$ indicates the model is ahead of the reference pace and $A_k<\ell_k$ indicates it is behind. The two one-sided signals are
$\mathrm{speed\_up} = \sum_{k=0}^{K-1} w^{\mathrm{up}}_k \pos{A_k - h_k} \ge 0$
and
$\mathrm{slow\_down} = \sum_{k=0}^{K-1} w^{\mathrm{dn}}_k \pos{\ell_k - A_k} \ge 0$
\begin{align}               
    u_\beta &= \mathrm{speed\_up} - \mathrm{slow\_down}
    \label{eq:u_beta}
\end{align}
where $\pos{z}=\max(0,z)$. The weights emphasize medium/hard buckets for
speeding up and easy buckets for slowing down. Pace is updated toward a
reference-anchored, clipped, EMA-smoothed target:
\begin{align}
    \beta_{\mathrm{target}}
      &= \clip\!\left(\beta_{\mathrm{ref}} - \eta_\beta\, u_\beta,\;
                      \beta_{\min},\,\beta_{\max}\right)
      \label{eq:beta_target}\\
    \beta_{t+1}
      &= (1-\lambda_\beta)\,\beta_t + \lambda_\beta\,\beta_{\mathrm{target}}
      \label{eq:beta_update}
\end{align}
Anchoring to $\beta_{\mathrm{ref}}$ rather than the current $\beta_t$ ensures
that when $u_\beta=0$, the target returns to $\beta_{\mathrm{ref}}$, allowing
the pace to relax toward the baseline schedule rather than remain at a
previously adapted value. Thus, $u_\beta>0$ accelerates progression, while
$u_\beta<0$ slows it.

\paragraph{Spread control via forgetting}
The controller tracks the best probe accuracy per bucket,
$A_k^{\mathrm{best}}=\max_{\tau\le t}A_k^{(\tau)}$, and defines forgetting as
$F_k=\pos{A_k^{\mathrm{best}}-A_k}$. Over designated lower-difficulty buckets
$\mathcal{E}\subseteq{0,\ldots,K-1}$,
\begin{equation}
u_\sigma
= c_{\mathrm{fgt}}\max_{k\in\mathcal{E}}F_k \ge 0
\label{eq:u_sigma}
\end{equation}
and the spread is updated toward a reference-anchored, clipped, EMA-smoothed target:
\begin{align}
    \sigma_{\mathrm{target}}
      &= \clip\!\left(\sigma_{\mathrm{ref}} + \eta_\sigma\, u_\sigma,\;
                      \sigma_{\min},\,\sigma_{\max}\right) \\
    \sigma_{t+1}
      &= (1-\lambda_\sigma)\,\sigma_t + \lambda_\sigma\,\sigma_{\mathrm{target}}
    \label{eq:sigma_update}
\end{align}
Because $u_\sigma\ge0$, spread control only broadens the sampling distribution
when easier buckets regress and relaxes toward $\sigma_{\mathrm{ref}}$ once
the regression clears. Both $\beta$ and $\sigma$ remain bounded around their
reference values to avoid degenerate curriculum
schedules. Algorithm~\ref{alg:ours} summarizes the complete training
procedure.
\begin{algorithm}[t]
\caption{\ours{}: closed-loop curriculum for GRPO}
\label{alg:ours}
\begin{algorithmic}[1]
\small
\Require difficulty buckets $\mathcal{D}_0,\ldots,\mathcal{D}_{K-1}$; budget $T$; probe
         interval $\Delta$; reference $(\beta_{\mathrm{ref}},\sigma_{\mathrm{ref}})$
\State $\beta_0 \gets \beta_{\mathrm{ref}},\quad \sigma_0 \gets \sigma_{\mathrm{ref}}$
\For{$t = 0$ \textbf{to} $T-1$}
  \State compute $x_t$ and $p_t(\cdot)$ from $(\beta_t,\sigma_t)$
         \Comment{Eqs.~\eqref{eq:adaptive_center}--\eqref{eq:adaptive_probs}}
  \State sample buckets $\sim p_t$, draw prompts, generate and score $G$ rollouts, update
         policy with GRPO
  \If{$t \bmod \Delta = 0$}
    \State $\Acc \gets$ read-only greedy probe over buckets
    \State $\beta_{t+1} \gets$ pace update$(\Acc)$
           \Comment{Eqs.~\eqref{eq:u_beta}--\eqref{eq:beta_update}}
    \State $\sigma_{t+1} \gets$ spread update$(\Acc)$
           \Comment{Eqs.~\eqref{eq:u_sigma}--\eqref{eq:sigma_update}}
  \Else
    \State $\beta_{t+1} \gets \beta_t,\quad \sigma_{t+1} \gets \sigma_t$
  \EndIf
\EndFor
\end{algorithmic}
\end{algorithm}

%% file: sections/experiment.tex

\section{Experimental Results and Analysis}
\label{sec:results}

\subsection{Experimental Setup}
\label{sec:setup}

\noindent\textbf{Model and Optimization.} All experiments
 use \texttt{meta-llama/Llama-3.2-3B-Instruct} with LoRA fine-tuning. LoRA
 uses rank $r=32$, scale $\alpha_{\mathrm{LoRA}}=64$, and dropout $0.05$.
 Training runs for $1{,}600$ optimizer steps with a learning rate of
 $10^{-5}$, cosine decay, and a $0.03$ warm-up ratio. The per-device batch
 size is $4$ with $4$ gradient-accumulation steps. GRPO samples $4$
 generations per prompt with KL coefficient $0.001$ and maximum gradient norm
 $1.0$.

\noindent\textbf{Generation and Probing.} The maximum prompt and completion
 lengths are $1{,}024$ and $192$ tokens, respectively. Training generations
 use temperature $0.7$ and top-$p=1.0$. Controller probes are performed every
 $\Delta=100$ optimizer steps using a fixed set of $100$ examples from each
 difficulty bucket and greedy decoding. Periodic probing introduces
 additional inference cost; in one representative run, \ours{} required 12.94
 h versus 11.49 h for Fixed E2H-G and 10.82 h for GRPO, although these
 single-run timings are indicative only.

\noindent\textbf{Controller Configuration.}
The controller uses reference values
$\beta_{\mathrm{ref}}=0.50$ and $\sigma_{\mathrm{ref}}=0.50$,
with probability floor $p_{\min}=0.01$. The controller variables are
bounded by $\beta\in[0.35,0.70]$ and
$\sigma\in[0.40,0.80]$. The update parameters are
$\eta_\beta=\eta_\sigma=0.50$,
$\lambda_\beta=\lambda_\sigma=0.30$, and
$c_{\mathrm{fgt}}=2.0$. For buckets
$\mathcal{D}_0,\ldots,\mathcal{D}_3$, the upper and lower
thresholds are
$h=(0.60,0.30,0.15,0.08)$ and
$\ell=(0.30,0.10,0.04,0.02)$, respectively. The corresponding
upward and downward weighting vectors are
$w^{\mathrm{up}}=(0.25,0.75,1.00,1.00)$ and
$w^{\mathrm{dn}}=(1.00,1.00,0.75,0.25)$. Forgetting is monitored
over $\mathcal{E}=\{0,1\}$, corresponding to buckets
$\mathcal{D}_0$ and $\mathcal{D}_1$.

\subsection{Accuracy Across Difficulty Levels}
\label{sec:difficulty}

Across four seeds, average exact-match accuracy is $42.4\% \pm 14.0\%$
for \ours{}, $35.0\% \pm 11.4\%$ for Fixed E2H-G, and $26.4\% \pm 21.6\%$ for
GRPO (Table~\ref{tab}). Across difficulty levels, CLOSEDLOOPGAUSSIAN achieves
mean accuracies of 82.7\%, 52.3\%, 23.2\%, and 11.5\% for $n=2,3,4,5$,
respectively, compared with 73.1\%, 40.8\%, 16.3\%, and 9.7\% for Fixed
E2H-G. \ours{} exceeds Fixed E2H-G on all four reported seeds, but the
across-seed dispersion is large relative to the mean gaps; with only $n=4$
seeds, the comparison does not support a statistical-significance claim and
is therefore interpreted descriptively. The largest mean gains over Fixed
E2H-G occur at difficulty $n=3$ ($52.3\%$ vs.\ $40.8\%$) and $n=4$ ($23.2\%$
vs.\ $16.3\%$), consistent with adaptation being most useful where rollouts
are neither uniformly correct nor uniformly incorrect. Each bucket contains
$250$ evaluation problems, so one example corresponds to $0.4$ percentage
points; small differences at $n=5$ therefore represent only a few examples
and should not be overinterpreted.

\begin{table}[t]
\centering
\caption{Per-seed average exact-match accuracy (\%). Mean and standard
deviation are computed across four seeds.}
\label{tab}
\begin{adjustbox}{max width=\linewidth}
\small
\begin{tabular}{lccccc}
\toprule
Method & 42 & 43 & 44 & 1 & Mean $\pm$ SD \\
\midrule
GRPO & 43.2 & 47.0 & 7.7 & 7.7 & $26.4 \pm 21.6$ \\
Fixed E2H-G & 44.5 & 41.9 & 19.2 & 34.3 & $35.0 \pm 11.4$ \\
\ours{} & \textbf{45.8} & \textbf{58.9} & \textbf{25.1} & \textbf{39.9}
& $\boldsymbol{42.4 \pm 14.0}$ \\
\bottomrule
\end{tabular}
\end{adjustbox}
\vspace{-17pt}
\end{table}

\subsection{Controller Dynamics: When Does Adaptation Help?}
\label{sec:control}
\begin{figure}[ht]
\centering
\includegraphics[width=\linewidth]{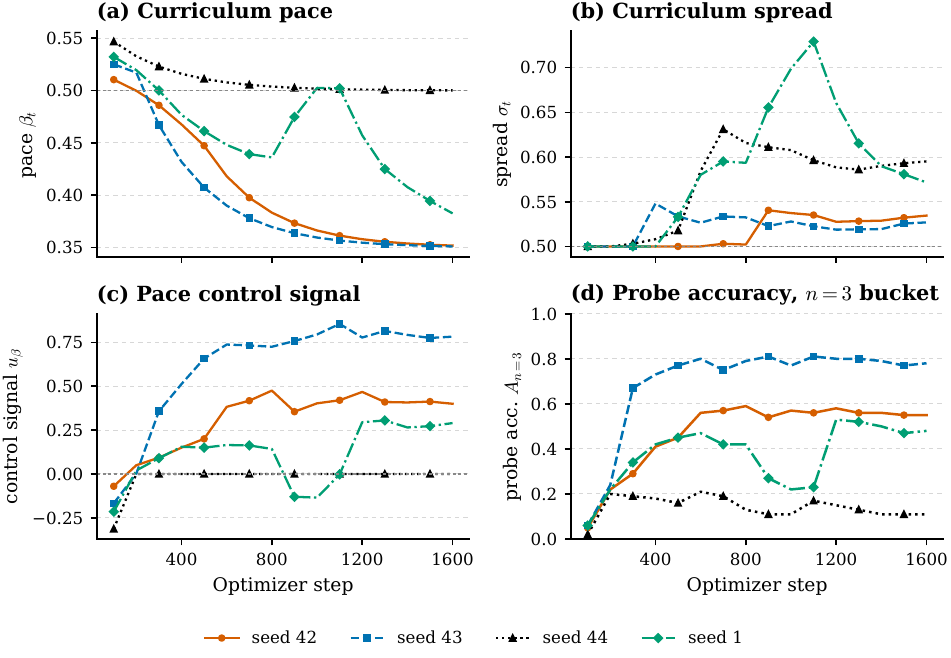}
\caption{Controller dynamics across the four seeds.
\textbf{(a)} pace $\beta_t$; \textbf{(b)} spread $\sigma_t$;
\textbf{(c)} pace control signal $u_\beta$;
\textbf{(d)} probe accuracy on the $n{=}3$ bucket.
Dashed grey lines mark the reference values ($\beta_{\mathrm{ref}}=\sigma_{\mathrm{ref}}=0.5$, $u_\beta=0$). Seeds $42$/$43$/$1$ move ahead of pace and the controller accelerates the curriculum ($\beta_t\!\downarrow$); seed $44$ never gets ahead, $u_\beta$ stays at zero, and the controller instead widens the distribution ($\sigma_t\!\uparrow$) to counter forgetting.}
\label{fig:dynamics}
\vspace{-10pt}
\end{figure}

Figure~\ref{fig:dynamics} traces $\beta_t$, $\sigma_t$, $u_\beta$, and $n=3$
probe accuracy across all four seeds. The first probe at step $100$ shows
near-floor accuracy beyond the easiest bucket, producing a mildly negative
$u_\beta$ and a small increase in $\beta_t$ above its $0.5$ reference. After
warm-up, two qualitative regimes emerge. On seeds $42$, $43$, and $1$, $n=3$
probe accuracy rises, $u_\beta$ becomes positive, and $\beta_t$ decreases
toward approximately $0.35$--$0.40$, shifting progression toward harder
buckets; $\sigma_t$ remains near its reference with only modest widening.
Seed $43$ shows the strongest acceleration and the highest final accuracy,
while seed $1$ exhibits a transient mid-training dip before returning to the
acceleration regime.

Seed $44$ follows a different trajectory: $n=3$ competence remains low,
$u_\beta$ returns to zero after step $200$, and $\beta_t$ relaxes near the
reference, while forgetting on easier buckets widens $\sigma_t$ to
approximately $0.63$. This near-reference $\beta_t$ is the intended
reference-anchoring behavior when no pace-adjustment signal is present. The
fixed controller therefore exhibits either accelerated progression or
increased rehearsal depending on the observed probe trajectory. As a
single-seed auxiliary analysis, seed $43$ sampler logs show that the
closed-loop schedule assigns $6.45$ percentage points more sampling mass to
$n=5$ than Fixed E2H-G, with the shift drawn primarily from $n=3$ and $n=4$.
Notably, $n=3$ accuracy improves despite receiving less $n=3$ training
exposure, suggesting that the gain cannot be explained simply by sampling
that bucket more frequently.

\subsection{Diagnostic Sensitivity to Controller Authority}
\label{sec:disgnostic}
Seed $43$ is examined because it exhibits the largest controller movement,
making sensitivity to controller authority easiest to observe. This seed was
selected after inspecting the controller trajectories, so the analysis is
post hoc and is reported as a diagnostic rather than a representative
ablation. On this seed, locking $\sigma=0.50$ reduces average accuracy from
$58.9$ to $26.2$, indicating that spread adaptation can matter for this
trajectory. Pace-bound changes also produce substantial variation, with the
default $[0.35,0.70]$ band outperforming both tighter and wider alternatives;
this band was fixed before the reported
runs (\S\ref{sec:closed_loop_controller}) and was not selected on the basis
of these evaluation results. Because this diagnostic uses one selected seed
and omits the complementary $\beta$-locked/$\sigma$-active condition, it does
not establish an average contribution of $\sigma$ or a causal decomposition
of pace and spread.

\subsection{Seed-Dependent Stability}
\label{sec:results_stability}
The large standard deviations in Table~\ref{tab} are consistent with training
instability in the underlying GRPO optimization rather than by the
curriculum. Vanilla GRPO is the least stable arm: it collapses to near-floor
accuracy on two of the four seeds ($44$ and $1$, both $7.7\%$) while reaching
$43$--$47\%$ on the other two, so its variance is bimodal rather than
Gaussian and its mean of $26.4\%$ should be read with that in mind. These two
seeds are separate runs with their own output directories and independently
written checkpoints; their identical per-bucket accuracies reflect two runs
reaching the same near-floor outcome, not a duplicated result entry. The
curriculum arms avoided near-floor outcomes on the same initializations:
Fixed E2H-G never falls below $19.2\%$, and \ours{} never below $25.1\%$. We
therefore treat seed $44$ not as an outlier to be averaged away but as a
documented stability observation, and seed $1$ shows the same effect for GRPO
a second time: closed-loop adaptation provides a measure of robustness to an
unfavorable optimization trajectory precisely because its pace and spread
react to the model's deteriorating competence. At the same time, we are
explicit that with four seeds on a single $3$B model and a single task, these
results give a \emph{consistent ordering and a mechanistic account}, not
a tight quantitative estimate of the expected gain.

%% file: sections/limitations.tex
\section{Limitations and Future Directions}
\label{sec:limitations}

This study focuses on a controlled setting: a single $3$B model (Llama-3.2-3B with LoRA), a single task (Countdown), and four seeds. 
The broader model, task, and seed coverage would further refine quantitative estimates of the expected gain, especially given the seed-to-seed variation observed in \S\ref{sec:results_stability}.  
Direct experimental comparisons with adaptive curricula such as AdaRFT, DUMP, and SEC are also left for future work. The controller's effectiveness also depends on the choice of thresholds and control gains. Controller thresholds, gains, and bounds are manually calibrated; overly strict settings can cause curriculum deadlock, although the reference-anchored bounded design mitigates this failure mode. Future work should expand seed and domain coverage, compare directly with adaptive-curriculum baselines, and automate controller calibration from training signals.

%% file: sections/conclusion.tex
\section{Conclusion}

This work studied when closed-loop curriculum adaptation improves reinforcement fine-tuning for LLM reasoning. We introduced \ours{}, a feedback-driven curriculum scheduler for GRPO that treats training-time data selection as an adaptive decision problem. Rather than changing the reward function, model architecture, or GRPO optimizer, \ours{} reallocates a fixed rollout budget across difficulty buckets using periodic probe estimates of per-bucket accuracy. It achieved higher mean exact-match accuracy than fixed E2H-G and vanilla GRPO across four reported seeds, although seed-to-seed variation is substantial. More broadly, this study highlights the value of decision-guided training mechanisms that improve reasoning by changing \emph{how} compute is allocated, rather than by increasing model size or redesigning the optimizer, establishing closed-loop curriculum scheduling as a lightweight and transparent component for building more efficient AI training systems.

%% file: sections/acknowledgment.tex
\section*{Acknowledgment}

This work is partially supported by the U.S. National Science Foundation under
Grants ITE-2431845 and IIS-2551684. Any opinions, findings, conclusions, or
recommendations expressed herein are those of the authors and do not
necessarily reflect the views of the U.S. NSF.

%% file: references.bib
@article{shao2024deepseekmath,
  title   = {DeepSeekMath: Pushing the Limits of Mathematical Reasoning in Open Language Models},
  author  = {Shao, Zhihong and Wang, Peiyi and Zhu, Qihao and Xu, Runxin and Song, Junxiao and Bi, Xiao and Zhang, Haowei and Zhang, Mingchuan and Li, Y. K. and Wu, Y. and others},
  journal = {arXiv preprint arXiv:2402.03300},
  year    = {2024}
}

@inproceedings{bengio2009curriculum,
title     = {Curriculum Learning},
author    = {Bengio, Yoshua and Louradour, J\'{e}r\^{o}me and Collobert, Ronan and Weston, Jason},
booktitle = {Proceedings of the 26th Annual International Conference on Machine Learning},
pages     = {41--48},
year      = {2009},
publisher = {Association for Computing Machinery},
doi       = {10.1145/1553374.1553380},
url       = {https://doi.org/10.1145/1553374.1553380}
}

@inproceedings{parashar2026e2hreasoner,
title={Curriculum Reinforcement Learning from Easy to Hard Tasks Improves {LLM} Reasoning},
author={Shubham Parashar and Shurui Gui and Xiner Li and Hongyi Ling and Sushil Vemuri and Blake Olson and Eric Li and Yu Zhang and James Caverlee and Dileep Kalathil and Shuiwang Ji},
booktitle={The Fourteenth International Conference on Learning Representations},
year={2026},
url={https://openreview.net/forum?id=KJvHnl3kUv}
}

@article{shi2025adarft,
  title={Efficient reinforcement finetuning via adaptive curriculum learning},
  author={Shi, Taiwei and Wu, Yiyang and Song, Linxin and Zhou, Tianyi and Zhao, Jieyu},
  journal={arXiv preprint arXiv:2504.05520},
  year={2025}
}

@article{yao2025sharegrpo,
  title   = {R1-ShareVL: Incentivizing Reasoning Capability of Multimodal Large Language Models via Share-GRPO},
  author  = {Yao, Huanjin and Yin, Qixiang and Zhang, Jingyi and Yang, Min and Wang, Yibo and Wu, Wenhao and Su, Fei and Shen, Li and Qiu, Minghui and Tao, Dacheng and Huang, Jiaxing},
  journal = {arXiv preprint arXiv:2505.16673},
  year    = {2025}
}

@article{guo2025deepseekr1,
  title={Deepseek-r1: Incentivizing reasoning capability in llms via reinforcement learning},
  author={Guo, Daya and Yang, Dejian and Zhang, Haowei and Song, Junxiao and Zhang, Ruoyu and Xu, Runxin and Zhu, Qihao and Ma, Shirong and Wang, Peiyi and Bi, Xiao and others},
  journal={arXiv preprint arXiv:2501.12948},
  year={2025}
}

@article{kimi2025k15,
  title={Kimi k1.5: Scaling reinforcement learning with llms},
  author={Kimi Team and Du, Angang and Gao, Bofei and Xing, Bowei and Jiang, Changjiu and Chen, Cheng and Li, Cheng and Xiao, Chenjun and Du, Chenzhuang and Liao, Chonghua and others},
  journal={arXiv preprint arXiv:2501.12599},
  year={2025}
}

@article{jaech2024openaio1,
  title={Openai o1 system card},
  author={Jaech, Aaron and Kalai, Adam and Lerer, Adam and Richardson, Adam and El-Kishky, Ahmed and Low, Aiden and Helyar, Alec and Madry, Aleksander and Beutel, Alex and Carney, Alex and others},
  journal={arXiv preprint arXiv:2412.16720},
  year={2024}
}

@article{yu2025dapo,
  title   = {{DAPO}: An Open-Source {LLM} Reinforcement Learning System at Scale},
  author  = {Yu, Qiying and Zhang, Zheng and Zhu, Ruofei and Yuan, Yufeng and Zuo, Xiaochen and Yue, Yu and Fan, Tiantian and Liu, Gaohong and Liu, Lingjun and Liu, Xin and others},
  journal = {arXiv preprint arXiv:2503.14476},
  year    = {2025}
}

@article{zheng2025gspo,
  title   = {Group Sequence Policy Optimization},
  author  = {Zheng, Chujie and Liu, Shixuan and Li, Mingze and Chen, Xiong-Hui and Yu, Bowen and Gao, Chang and Dang, Kai and Liu, Yuqiong and Men, Rui and Yang, An and others},
  journal = {arXiv preprint arXiv:2507.18071},
  year    = {2025}
}

@article{narvekar2020curriculum,
  title   = {Curriculum Learning for Reinforcement Learning Domains: A Framework and Survey},
  author  = {Narvekar, Sanmit and Peng, Bei and Leonetti, Matteo and Sinapov, Jivko and Taylor, Matthew E. and Stone, Peter},
  journal = {Journal of Machine Learning Research},
  volume  = {21},
  number  = {181},
  pages   = {1--50},
  year    = {2020}
}

@inproceedings{graves2017automated,
  title     = {Automated Curriculum Learning for Neural Networks},
  author    = {Graves, Alex and Bellemare, Marc G. and Menick, Jacob and Munos, R{\'e}mi and Kavukcuoglu, Koray},
  booktitle = {Proceedings of the 34th International Conference on Machine Learning (ICML)},
  pages     = {1311--1320},
  year      = {2017}
}

@article{matiisen2020teacher,
  title   = {Teacher-Student Curriculum Learning},
  author  = {Matiisen, Tambet and Oliver, Avital and Cohen, Taco and Schulman, John},
  journal = {IEEE Transactions on Neural Networks and Learning Systems},
  volume  = {31},
  number  = {9},
  pages   = {3732--3740},
  year    = {2020}
}

@INPROCEEDINGS{chen2025generativejudge,
  author={Chen, Changxin},
  booktitle={2025 IEEE 37th International Conference on Tools with Artificial Intelligence (ICTAI)},
  title={Training Generative Judge with Hard Negative Mining: A Metric Learning Perspective},
  year={2025},
  pages={778-785},
  doi={10.1109/ICTAI66417.2025.00112}}

@article{bae2025onlinedifficulty,
  author={Bae, Sanghwan and Hong, Jiwoo and Lee, Min Young and Kim, Hanbyul and Nam, JeongYeon and Kwak, Donghyun},
  title={Online Difficulty Filtering for Reasoning Oriented Reinforcement Learning},
  journal={arXiv preprint arXiv:2504.03380},
  year={2025}}

@article{wang2025dump,
  author={Wang, Zhenting and Cui, Guofeng and Wan, Kun and Zhao, Wentian},
  title={{DUMP}: Automated Distribution-Level Curriculum Learning for {RL}-based {LLM} Post-training},
  journal={arXiv preprint arXiv:2504.09710},
  year={2025}}

@article{chen2025selfevolving,
  author={Chen, Xiaoyin and Lu, Jiarui and Kim, Minsu and Zhang, Dinghuai and Tang, Jian and Pich{\'e}, Alexandre and Gontier, Nicolas and Bengio, Yoshua and Kamalloo, Ehsan},
  title={Self-Evolving Curriculum for {LLM} Reasoning},
  journal={arXiv preprint arXiv:2505.14970},
  year={2025}}
